# Enhanced Semantic Segmentation Pipeline for WeatherProof Dataset Challenge

Nan Zhang, Xidan Zhang, Jianing Wei, Fangjun Wang, Zhiming Tan
Fujitsu R&D Center Co., Ltd.
{zhangnan, zhangxidan, weijianing, wangfangjun, zhmtan}@fujitsu.com

## Abstract

*This report describes the winning solution to the WeatherProof Dataset Challenge (CVPR 2024 UG2+ Track 3). Details regarding the challenge are available at https://cvpr2024ug2challenge.github.io/track3.html. We propose an enhanced semantic segmentation pipeline for this challenge. Firstly, we improve semantic segmentation models, using backbone pretrained with Depth Anything to improve UperNet model and SETRMLA model, and adding language guidance based on both weather and category information to InternImage model. Secondly, we introduce a new dataset WeatherProofExtra with wider viewing angle and employ data augmentation methods, including adverse weather and super-resolution. Finally, effective training strategies and ensemble method are applied to improve final performance further. Our solution is ranked 1$^{st}$ on the final leaderboard. Code will be available at [https://github.com/KaneiGi/WeatherProofChallenge](https://github.com/KaneiGi/WeatherProofChallenge).*

## 1. Introduction

Semantic Segmentation has a long history of success in various fields, but recent advancements hinge on the use of large foundational models. These models excel on benchmarks like ADE20K and Cityscapes, but their performance suffers significantly when encountering images with visual degradation, such as those taken in bad weather. The WeatherProof Dataset Challenge (CVPR 2024 UG2+ Track2) introduces the first collection of accurately paired clear and weather-degraded image pairs. This challenge tackles real-world weather effects on semantic segmentation, aiming to spark new methods for handling these challenging images.

In this technical report, we provide a brief introduction to the winning solution to the WeatherProof Dataset Challenge, see [1]. To address the problems of semantic segmentation in adverse weather, including poor image quality, rain and snow noise interference, and large scene differences, we first propose stronger semantic segmentation models. And then we introduce a new dataset and employ data augmentation methods. Finally, effective training strategies and ensemble method are applied to improve final performance.

## 2. Approach

### 2.1. Semantic segmentation models

#### 2.1.1. Semantic segmentation models based on Depth Anything

There are two methods using Depth Anything [2] pre-trained backbone in our solution. Depth Anything is a MDE (Monocular Depth Estimation) model. It inherits the rich semantic priors from a pre-trained DINOv2 via a simple feature alignment constraint. In the other hand, it has been trained on a large-scale unlabeled dataset and has good generalization ability, so we choose the Depth Anything pre-trained model as the backbone.

One method using Depth Anything is based on UperNet model. The backbone is DINOV2 pretrained with Depth Anything. It uses Feature2Pyramid as neck, UPerHead [3] as Decoder_head and SegmenterMaskTransformerHead [4] as Auxiliary_head. Decoder_head loss and Auxiliary_head loss are both CrossEntropyLoss. The key components of model architecture are shown in Figure 1.

Another method is based on SETRMLA model. The backbone is also DINOV2 pretrained with Depth Anything. It uses MLANeck [5], SETRMLA [5] as Decoder_head and four FCNHead [6] as Auxiliary_heads. Decoder_head loss and Auxiliary_head loss are both CrossEntropyLoss.

#### 2.1.2 Semantic segmentation model based on language guidance

The original WeatherProof model utilizes weather condition prompts to guide the semantic segmentation [7, 8, 9]. To enhance the model's performance, we do two improvements, enriching the prompt words and enhancing CLIP image input methodology.

We first introduce additional category prompts to CLIP. By incorporating both weather and object category information, the model gains a more comprehensive understanding of the scene, leading to improved segmentation accuracy. This allows the model to differentiate between various objects more effectively, even under adverse weather conditions.

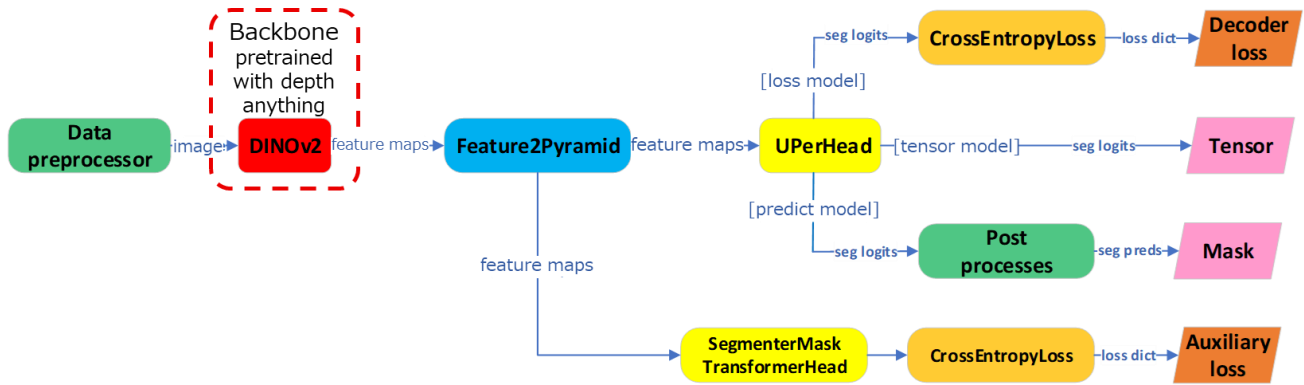

Figure 1: Key components of UperNet model with depth anything pretrained backbone

The original CLIP model used by WeatherProof can only process images at a resolution of 225x225 pixels. This requires resizing the entire image, which can lead to loss of critical information. To address this, we implement a sliding window approach. By dividing the image into smaller patches and processing each patch individually through CLIP, we retain higher resolution details. The final segmentation map is generated by averaging the results from all patches, leading to a more accurate and detailed output.

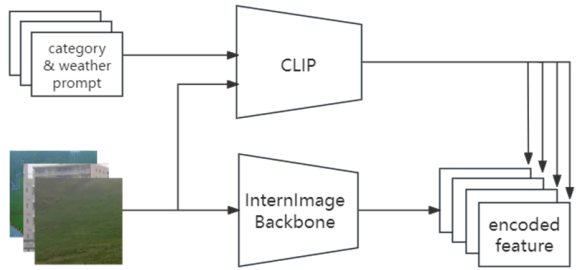

Figure 2: Overview of segmentation model based on CLIP guidance. CLIP injections are added to the end of each stage.

### 2.1.3 Other semantic segmentation models

Except for above improved semantic segmentation models, we also try other models, and choose following two for our solution. The first one is OneFormer model [10], a multi-task universal image segmentation framework based on transformers. In this work, we use text "the task is semantic" to encode task token, semantic text to encode text token, and use large scale Swin-Transformer as backbone. The second is InternImage model [9], which uses deformable convolutions to maintain the long-range dependence of attention layers in a low memory/computation regime. We use InternImage's XL backbone and keep its 39 layers in 4 stages.

### 2.2. Datasets

We use three datasets for training: WeatherProof, WeatherProofClean, and WeatherProofExtra. WeatherProof is the original dataset provided by the challenge organizer, and it contains 513 scenes for training, 38 scenes for validation. During testing phase, all the data is used for training. WeatherProofClean has the clean images for each scene in WeatherProof, and the annotations are the same. WeatherProofExtra is an extra dataset, and it can be downloaded from huggingface repository[1]. It has 160 scenes, including rainy, snowy and foggy.

We do an augmentation on the extra data, including adverse weather and super-resolution. Figure 3 shows an example of the augmented images. We use rainy, snowy or lighting weather watermarks to cover the original image to simulate various adverse weather conditions. Also, by comparing the images in testing dataset with previous training dataset, we find that the resolution of testing images is higher than the training ones. So, we use super-resolution method [11] to narrow the gap between training and testing resolutions.

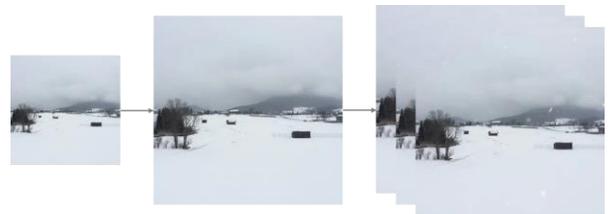

Figure 3: Example of augmentation for extra data. From left to right: original image, enlarged image by super-resolution, images covered with weather-marks.

---

[1]https://huggingface.co/datasets/WangFangjun/WeatherProofExtra

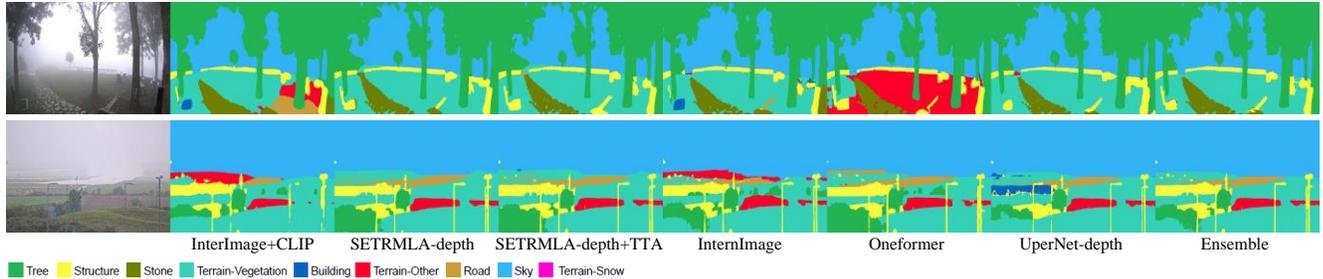

Figure 4: Visual comparison between different models and final ensemble result.

## 2.3. Training strategy

We first train the models using the original datasets WeatherProof and WeatherProofClean. Basic data enhancement operations, including image flipping, cropping, rotation, scaling, etc., are also performed on the training set. In addition, after analyzing the training, validation and test sets, we identify that the test sets have higher resolution and wider viewing angle. Then we use WeatherProofExtra dataset to finetune the models with larger image size as training input.

## 3. Experiments

### 3.1. Training details

We train five models separately in different training details. For InternImage and InternImage + CLIP models, the input size of training is set to 640*640 pixels. We use the AdamW optimizer with an initial learning rate of 0.00002 and a weight decay of 0.05. The model is trained for a total of 10 epochs, with a batch size of 4. For Depth Anything based models, SETRMLA-depth and UperNet-depth, the input images are randomly scaled from 448 to 1882 on short side and up to 3584 on the long side, then randomly cropped and padded to 896*896. We use AdamW optimizer with an initial learning rate of 0.00003. The models are trained for 160K iterations with batch size 4. Data augmentation techniques include random flip and random crop both with a probability of 0.5, and PhotoMetricDistortion. For Oneformer model, the image size during training is set to a randomly picked value from [960, 992, 1024] for each iteration, and set 1024 for inference. The base learning rate is set to 1e-4, and the model is fine-tuned with 7500 iterations.

### 3.2. Ablation study

For the model with language guidance mentioned in 2.1.2, to evaluate the effectiveness of our prompt word, we conduct an ablation study. We take InternImage-XL as the basic model, then compare the result of adding category prompts and weather prompts. All methods use an input size of 512*512, trained and tested on WeatherProof train & evaluate split. The results are summarized in Table 1.

Table 1: Ablation study result from language guidance

| Method | mIoU |
|---|---|
| Category Prompts Only | 0.534 |
| Weather Prompts Only | 0.535 |
| Category + Weather Prompts | 0.552 |

We further evaluate various strategies using the WeatherProof Dataset Challenge test set. All methods use an input size of 640*640. The baseline is InternImage-XL model. CLIP Guidance/C method means our improved model by adding language guidance mentioned in 2.1.2. Finetuning means the training strategy mentioned in 2.3. Inference enhancement involves sliding window operation and scaling images from 0.1 to 1.5 times, then calculating the average. The results are shown in Table 2.

Table 2: Ablation study experiments result

| Method | mIoU |
|---|---|
| Baseline | 0.395 |
| + CLIP Guidance | 0.402 |
| + C + Finetuning | 0.430 |
| + C + F + Inference Enhancement | 0.439 |

### 3.3. Experimental results

We evaluate the performance of different methods, and finally we choose six results of five models to ensemble. The ensemble method is to vote for each pixel in six predicted results and select the category with the highest number of votes as the final category.

The results are listed in Table 3. For single model, SETRMLA-depth can reach the mIoU of 0.46. After ensemble for the six results, our final mIoU is 0.47.

Table 3: Experimental results

| Method | mIoU |
|---|---|
| InternImage | 0.43 |
| InternImage + CLIP | 0.44 |
| UperNet-depth | 0.42 |
| **SETRMLA-depth** | **0.46** |
| SETRMLA-depth + TTA | 0.43 |
| Oneformer | 0.43 |
| **Ensemble** | **0.47** |

## 4. Conclusions

In this report, we describe the winning solution of WeatherProof Dataset Challenge. To tackle the problems of semantic segmentation in adverse weather, we employ strong segmentation models and make improvements. We also provide an extra dataset to narrow the gap between training and testing data. Finally effective training strategies and ensemble method are applied. Our result shows that model with backbone pretrained by Depth Anything has the best mIoU as a single model. After ensemble of all the results, our solution can get final mIoU of 0.47.